\documentclass[10pt,twocolumn,letterpaper]{article}

\usepackage[final]{cvpr}      % To produce the REVIEW version
\usepackage[table]{xcolor}
\usepackage{xcolor}
\usepackage{xcolor}
\usepackage{multirow}
\definecolor{SteelBlue}{RGB}{70,130,180}
\newcommand{\tablestyle}[2]{%
  \setlength{\tabcolsep}{#1}%
  \renewcommand{\arraystretch}{#2}%
  \centering
}

\definecolor{cvprblue}{rgb}{0.21,0.49,0.74}
\usepackage[pagebackref,breaklinks,colorlinks,allcolors=cvprblue]{hyperref}

\def\paperID{14103} % *** Enter the Paper ID here
\def\confName{CVPR}
\def\confYear{2026}

\title{CoMo: Learning Continuous Latent Motion from Internet Videos \\ for Scalable Robot Learning}

\usepackage{bbding} % 确保引入了图标包
\makeatletter
\renewcommand*{\@fnsymbol}[1]{\ensuremath{\ifcase#1\or \mbox{\Envelope}\or \dagger\or \ddagger\or \mathsection\or \mathparagraph\or \|\or **\or \dagger\dagger \or \ddagger\ddagger \else\@ctrerr\fi}}
\makeatother

\author{Jiange Yang$^{1}$ \quad Yansong Shi$^{2,3}$ \quad Haoyi Zhu$^{2,3}$ \quad Mingyu Liu$^{2,4}$ \\ Kaijing Ma$^{2,5}$ \quad Yating Wang$^{2,6}$  \quad Gangshan Wu$^{1}$ \quad Tong He$^{2}$ \quad Limin Wang$^{1, 2,}$\thanks{Corresponding author.}\\
$^1$Nanjing University, $^2$Shanghai AI Lab, 
$^3$University of Science and Technology of China \\
$^4$Zhejiang University, $^5$Fudan University,  $^6$Tongji University \\
\tt\small jiangeyang.jgy@gmail.com, lmwang@nju.edu.cn\\ 
}

\begin{document}
\maketitle
\begin{abstract}

% However, existing discrete methods suffer from information loss and struggle with complex and fine-grained dynamics. To mitigate the shortcut learning problem caused by extracting excessive static background information from consecutive latent motions, 

% Simultaneously, it utilizes the temporally reversed latent motion as a negative sample for contrastive learning.

Unsupervised learning of latent motion from Internet videos is crucial for robot learning. Existing discrete methods generally mitigate the shortcut learning caused by extracting excessive static backgrounds through vector quantization with a small codebook size. However, they suffer from information loss and struggle to capture more complex and fine-grained dynamics. Moreover, there is an inherent gap between the distribution of discrete latent motion and continuous robot action, which hinders the joint learning of a unified policy. We propose CoMo, which aims to learn more precise continuous latent motion from internet-scale videos. CoMo employs an early temporal difference (Td) mechanism to increase the shortcut learning difficulty and explicitly enhance motion cues. Additionally, to ensure latent motion better captures meaningful foregrounds, we further propose a temporal contrastive learning (Tcl) scheme. Specifically, positive pairs are constructed with a small future frame temporal offset, while negative pairs are formed by directly reversing the temporal direction. The proposed Td and Tcl work synergistically and effectively ensure that the latent motion focuses better on the foreground and reinforces motion cues. Critically, CoMo exhibits strong zero-shot generalization, enabling it to generate effective pseudo action labels for unseen videos. Extensive simulated and real-world experiments show that policies co-trained with CoMo pseudo action labels achieve superior performance with both diffusion and auto-regressive architectures. The code will be available at \textbf{\normalsize\url{https://github.com/MCG-NJU/CoMo}}.

% The shared continuous distribution of robot action and latent motion also benefits the joint learning of unified policy. 

\end{abstract}

\vspace{-1.5mm}    
\section{Introduction}
\label{sec:intro}

While large-scale Internet data has enabled impressive generalization in vision and language models~\citep{gpt4,sam}, robot learning remains limited by data scarcity, low diversity, and high heterogeneity. To enable effective scaling in robot learning, a recent popular paradigm~\citep{lapas,moto,igor,agibot,gr00t} focuses on learning latent motion models from extensive videos. They typically utilize an inverse dynamics encoder–forward dynamics decoder architecture within a self-supervised reconstruct objective using video frame pairs, simultaneously employing a VQ-VAE objective~\citep{vq} to quantize learned motion representations, to generate pseudo action labels for the unlabeled video data.

A key reason previous works favored discrete latent motion, using a small codebook (e.g., 8 in LAPA~\citep{lapas} and 16 in UniVLA~\citep{univla}), was to inhibit the risk of model collapse. When attempting to learn latent motion directly, models are highly susceptible to `shortcut learning'. Specifically, the inverse dynamics encoder may capture excessive static background information from the future frame, rather than strictly focusing on the foreground motion, simply because the decoder can then reconstruct pixel-level details more easily. This degeneracy turns the model into an ineffective future-frame predictor, severely subverting its utility as an action prediction mechanism suitable for co-training unified robot policies. Therefore, the highly quantized latent motion (with a small codebook) is employed to mitigate this issue by abstracting only the rough direction or macro-trend of the movement over a temporal horizon.

However, real-world motion is inherently continuous, characterized by complex and fine-grained dynamics. Representing motion with discrete codebook inevitably leads to information loss and limits generalization to novel motion patterns. Evidence from visual generation~\citep{mar,fluid} and robot learning~\citep{vla_improve,pi5} also suggests that continuous representations can yield superior performance. This motivates a question: \textit{Could we learn continuous latent motion from action-less videos?} Continuous latent motion enables more accurate representation of fine-grained inter-frame changes and inherently provides better consistency with the continuity of robot action to co-train unified policy.

To address the shortcut learning issue of directly learning continuous latent motion, our CoMo, firstly introduces a early temporal difference (Td) strategy, inspired by temporal difference networks~\citep{tdn} in video understanding. Specifically, we remove the direct encoding of future frames and replace it with feature differences between current and future frame before the encoder input, which serves to enhance dynamic motion cues while increasing the difficulty of shortcut learning. The sparse feature differences explicitly amplifies motion signals, but inevitably contains irrelevant information when the motion dimension is larger. To address this, our CoMo further incorporates a temporal contrastive learning mechanism: motion representations with a slight temporal offset in the future frames are treated as positive pairs, while those with reversed temporal direction between the current and future frames are used as negative pairs. The InfoNCE~\cite{infonce} loss is then applied to encourage effective discrimination to learn more structured representation. However, standalone temporal contrastive learning risks only capturing future-frame foreground object identity (‘what’ and ‘where’) while neglecting temporal motion patterns (‘how’). Therefore, the early temporal difference mechanism and temporal contrastive learning work synergistically to ensure that the continuous motion representations learned by CoMo are both focused on meaningful foreground regions and enriched with action-relevant motion cues. Consequently, CoMo leads to more accurate pseudo action labels for action-less video data.

An extra challenge in latent motion learning is the lack of direct, low-cost, and robust quantitative analysis tools to evaluate and analyze the learned motion representations without policy. We adopt two metrics for this purpose: (1) Action Prediction MSE (MSE). Given the robot dataset and following~\cite{villa,clam,supervision}, we train an MLP to regress ground-truth actions from the latent motion embeddings and report the MSE. It assesses the latent motion's encoding capacity for action-relevant information. (2) Past-to-Current and Future-to-Current Similarity (S-PCFC). Computed on motion-centric (i.e., data with minimized background variations) robot demonstration datasets, S-PCFC measures the cosine similarity between $z(O_{t-n}, O_t)$ and $z(O_{t+n}, O_t)$. Since high-dimensional latent motion inevitably introduces redundant, action-irrelevant noise such as future frame backgrounds, S-PCFC is designed to diagnose and quantify this action-irrelevant noise. Empirically, we find that the combination of MSE and S-PCFC effectively reflects the policy success rate, achieving the best policy performance when both metrics are relatively low.

Overall, CoMo can generate more effective pseudo action labels for action-less video data. The consistent continuous distribution between robot action and video latent motion directly facilitates unified and joint policy learning, removing complex multi-stage pretraining and finetuning procedures (~\citep{lapas}) or explicit two-stage motion-before-action pipelines (~\citep{agibot}). Finally, extensive simulation and real-world experiments validate that CoMo provides more precise, effective pseudo action labels and achieves superior policy performance compared to those using discrete latent motion or naive continuous baseline. In summary, our main contributions are as follows:

\begin{itemize}[left=0pt, itemsep=0.5pt]
  \item We propose CoMo, for unsupervised learning of more fine-grained, informative and continuous latent motion representations from Internet videos.
  \item Our CoMo removes vector quantization and proposes early temporal difference mechanism and temporal contrastive learning to collaboratively ensure that the continuous latent motion focuses more on meaningful foreground regions and enhances action-relevant motion cues.
  \item We generate more precise pseudo action labels using CoMo for action-less videos. The consistently continuous distributions of latent motion and robot action naturally facilitate the joint learning of unified policy.
  \item Extensive simulation and real-world experiments demonstrate the superior performance of CoMo across both diffusion and auto-regressive policies.
\end{itemize}

\vspace{-1mm}

\section{Related Work}
\label{sec:releated}

 \noindent \textbf{Learning from Internet Data for Robotic Manipulation.} Limited robot data restricts scalable policy learning. Incorporating large-scale internet videos can enhance generalization and data efficiency~\citep{survey}. Prevailing methods predict signals from video data, either as implicit auxiliary tasks for improved learning~\citep{gr2} or as explicit guidance for policy execution~\citep{atm}. These signals include future visual observations~\citep{STP,yilundu,gr2,gen2act,team2025aether,deepverse,dreamvla,worldvla,cosmospolicy}, affordances~\citep{rt-affor,cmu_affordance}, object masks~\citep{mask,mba}, optical flow~\citep{avdc}, human hand poses~\citep{swm,mimicplay,egovla,being05}, and sparse point trajectories~\citep{RT-Trajectory,flow,track2act,tramoe}. A recent popular framework~\citep{lapas,moto,igor,agibot,gr00t} utilizes an inverse dynamics encoder–forward dynamics decoder architecture with unsupervised VQ-VAE~\citep{vq} objectives to extract discrete latent motion from action-less videos, a choice often made to combat shortcut learning. Additionally, ~\citep{clam,supervision} introduce extra robot action supervision to train continuous latent motion IDM and ~\citep{adaworld,dreamdojo} learn continuous latent motion world model. In contrast, CoMo focuses on learning purely self-supervised, continuous latent motion IDM, avoiding the representational trade-offs inherent in VQ-VAE based methods.

\begin{figure*}[htbp]
\centering
\includegraphics[width=0.945\textwidth]{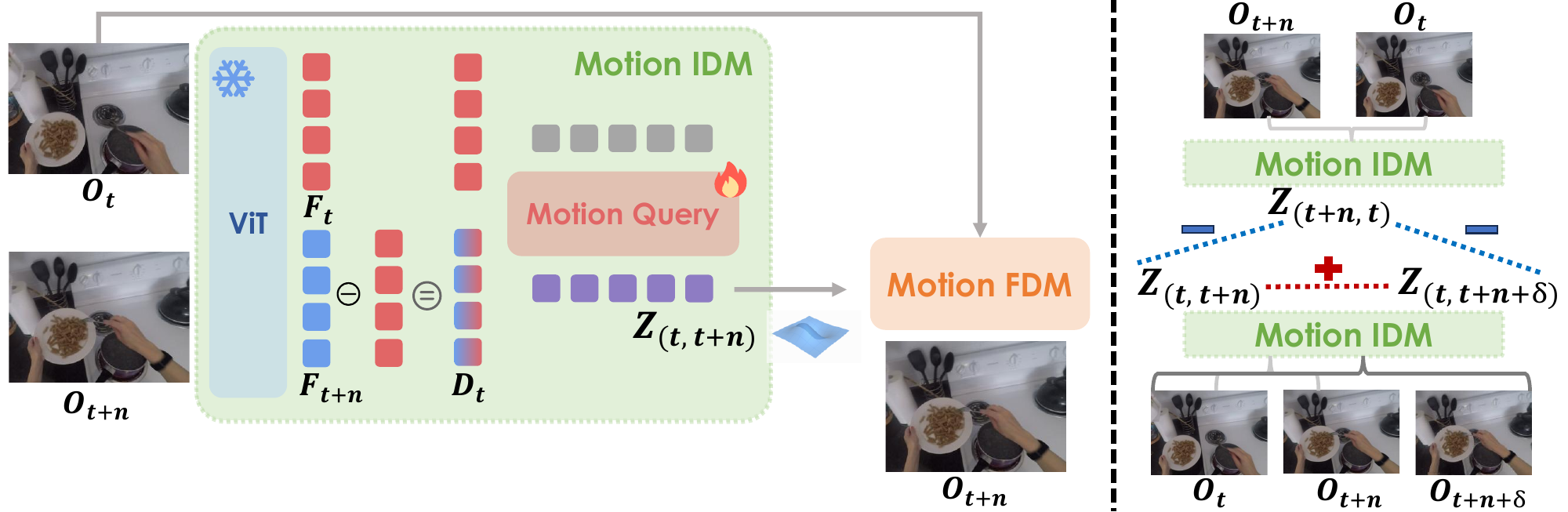} % Reduce the figure size so that it is slightly narrower than the column.
\caption{ \textbf{The CoMo framework.} \textbf{(Left)} CoMo model architecture. \textbf{(Right)} Temporal contrastive learning scheme. Built upon the standard IDM-FDM architecture, CoMo learn VQ-free, continuous and more precise inter-frame latent motion. CoMo introduces early temporal difference mechanism and temporal contrastive learning method to collaboratively ensure that the continuous latent motion focuses more on meaningful foreground regions and enhances action-relevant motion cues.}
\vspace{-4mm}
\label{method}
\end{figure*}

% Ultimately, CoMo's latent motion provides more accurate pseudo action labels for action-less video data, enabling seamless policy co-training with the naturally continuous distribution of robot action.

\noindent \textbf{Robotic Manipulation Policy Architecture.} Early works focused on state-based reinforcement learning~\citep{RL1,RL2}, while more recent methods leverage raw visual observations for imitation learning~\citep{bcz,bc2,fang2023rh20t,tpm}. Many methods use generative probabilistic modeling~\citep{rt1,diffusion-policy,act} to capture the complex multi-modal action distribution. Among these, some methods~\citep{rt2,openx,openvla,fast,vqvla,fact} adopt autoregressive-based policy architectures, which benefit from the generalization of pretrained VLMs but require discretizing robot actions.  In contrast, others~\citep{diffusion-policy, rdt,pi5} use diffusion-based policy architecture to generate continuous robot action directly. Recent studies~\citep{vla_improve, pi5} have shown that continuous action representations enable finer-grained behavior modeling and often yield better performance. As a result, many advanced approaches~\citep{tinyvla,cogact,pi0,hybridvla,pi5,ki} combine autoregressive VLM backbones with diffusion-based action experts, thereby benefiting from both the strong generalization ability of pretrained VLMs and the expressive power of continuous action representations. On this basis, CoMo could leverage a unified policy architecture to jointly learn both continuous robot action and video latent motion.

% \section{Formatting your paper}
% \label{sec:formatting}

\vspace{-1mm}

\section{Method}

\subsection{Learning Continuous Latent Motion}

We first describe our CoMo framework. CoMo adopts a inverse dynamics encoder–forward dynamics decoder paradigm, as illustrated on Fig.~\ref{method}. 

% Subsequently, we detail the technical aspects of our motion-enhanced inverse dynamics encoder and forward dynamics decoder respectively.

\noindent \textbf{Inverse dynamics Model (IDM) with early temporal difference (Td) mechanism.} Our IDM aims to extract precise and background-irrelevant continuous motion information. Given a pair of the current frame $O_t$ and the future frame $O_{t+n}$, we use a shared MAE~\citep{mae} pretrained ViT~\citep{vit} to extract their respective token-level features $F_t$ and $F_{t+n}$. Subsequently, to enhance action-relevant motion cues, we perform a early temporal difference operation by element-wise subtraction between $F_t$ and $F_{t+n}$ to obtain the token-level temporal feature differences $D_t$. To further suppress static backgrounds and shortcut learning, we explicitly remove the future frame features $F_{t+n}$ before the encoder extracting motion embeddings. Specifically, we concatenate only the current frame features $F_t$ and the temporal feature differences $D_t$, resulting in the combined representation $[F_t, D_t]$. Following Moto-GPT~\citep{moto}, we concatenate a set of learnable query embeddings with token-level combined representation and perform full attention interaction within standard multi-layer Transformer blocks (Motion Q-former). We then take the query features from the output of the Transformer layers as our motion representation $Z_{t, t+n}$.

\noindent \textbf{Latent motion learning with temporal contrastive learning (Tcl) scheme.} Extracting excessive static future frame background information poses a core challenge for learning continuous latent motion without vector quantization. In IDM, we enhance motion cues through temporal difference and increase the difficulty of shortcut learning by removing future frame features. However, sparse feature differences alone may contain substantial irrelevant information. To address this, we further propose a temporal contrastive learning approach. Specifically, we sample $O_{t+n+\delta}$ near $O_{t+n}$ and use IDM to obtain $Z_{t, t+n+\delta}$, as well as $Z_{t+n, t}$, and minimize the InfoNCE loss:

\vspace{-2mm}

\begin{align}
S_1 &= S(Z_{t, t+n+\delta},\ Z_{t, t+n}) \\
S_2 &= S(Z_{t, t+n+\delta},\ Z_{t+n, t})\\
S_3 &= S(Z_{t, t+n},\ Z_{t+n, t}) \\
\mathcal{L}_{\mathrm{tcl}} &= -\log \frac{e^{S_1}}{e^{S_1} + e^{S_2} + e^{S_3}}
\end{align}

\vspace{-2mm}

Among them, $S$ denotes the cosine similarity measure. The slightly temporally-offset $(Z_{t, t+n+\delta},\ Z_{t, t+n})$ forms the positive motion pair, while $(Z_{t, t+n+\delta},\ Z_{t+n, t})$ and $(Z_{t, t+n},\ Z_{t+n, t})$ are negative motion pairs with reversed temporal direction. In practice, we set the range of $\delta$ to $[-n/5, n/5]$. However, standalone temporal contrastive learning risks only capturing future-frame foreground object identity (‘what’ and ‘where’) while neglecting temporal motion patterns (‘how’). Therefore, the early temporal difference mechanism and temporal contrastive learning collaboratively mitigate the shortcut learning problem. Together, they ensure that the continuous latent motion learned by CoMo focuses more on meaningful foreground regions and enhances action-relevant motion cues. Moreover, CoMo does not require strong vector quantization constraints (such as a small codebook size or low-dimensional motion embeddings), making it more scalable.

\noindent \textbf{Forward dynamics Model (FDM).} Conditioned on the latent motion representation $Z_{t,t+n}$ obtained from the IDM and the current frame visual observation $O_t$, the forward dynamics decoder aims to reconstruct the the future frame observation $O_{t+n}$. Specifically, we first obtain a low-level patch-level embedding of $O_t$ using a linear patch embedding layer. Simultaneously, we further perform a pooling operation to compress the motion representation $Z_t$, and then add the pooled motion feature to the low-level patch-level embedding of $O_t$, resulting in $E (O_t, Z_{t,t+n})$. Subsequently, several Transformer layers process the combined features. Finally, the output features are processed using convolutional layers and a pixel shuffling operation to reconstruct the predicted future frame $\hat{O}_{t+n}$.

Finally, our CoMo framework is trained by jointly minimizing a weighted InfoNCE loss, a reconstruction loss and a perceptual loss to ensure both pixel-level accuracy and perceptual fidelity of the predicted future frames.

\subsection{Joint Unified Policy Learning}

% \vspace{-2mm}

We perform joint learning of action-less video data and continuous robot action data within a unified policy model. Specifically, given an action-labeled robot dataset $\boldsymbol{\mathcal{D}_{R}} = \{\tau_1, . . . , \tau_n\}$, where each $\tau_i$ represents a trajectory consisting of paired robot observations and actions, denoted as $\tau_i = [(o_0, a_0), \ldots, (o_T, a_T)]$, and a larger-scale action-less video data $\boldsymbol{\mathcal{D}_{V}}$, we utilize the trained CoMo IDM to extract continuous latent motion embeddings for $\boldsymbol{\mathcal{D}_{V}}$. As a result, each trajectory in $\boldsymbol{\mathcal{D}_{V}}$ can be augmented as $[(o_0, z_0), \ldots, (o_T, z_T)]$, where $z_t$ denotes the latent motion inferred by the IDM at time step $t$. Since both $a$ and $z$ exhibit continuous data distribution, we can seamlessly perform joint imitation learning using the combined dataset $\boldsymbol{\mathcal{D}_{R}} \boldsymbol{\cup} \boldsymbol{\mathcal{D}_{V}}$ within a unified generative policy. We simply need to allocate separate lightweight heads for $\boldsymbol{\mathcal{D}_{R}}$ and $\boldsymbol{\mathcal{D}_{V}}$. This flexible co-training strategy does not require complex multi-stage pretraining and finetuning procedures (~\cite{lapas}) or explicit two-stage motion-before-action pipelines (~\cite{agibot}). The ability to leverage larger-scale dataset sources allows our CoMo to offer a scalable robot learning paradigm. In this work, we develop both a unified diffusion-based policy and an auto-regressive policy.

\vspace{-1mm}

\section{Experiments}

\begin{figure}[htbp]
\label{vis}
\centering
\includegraphics[width=0.499\textwidth]{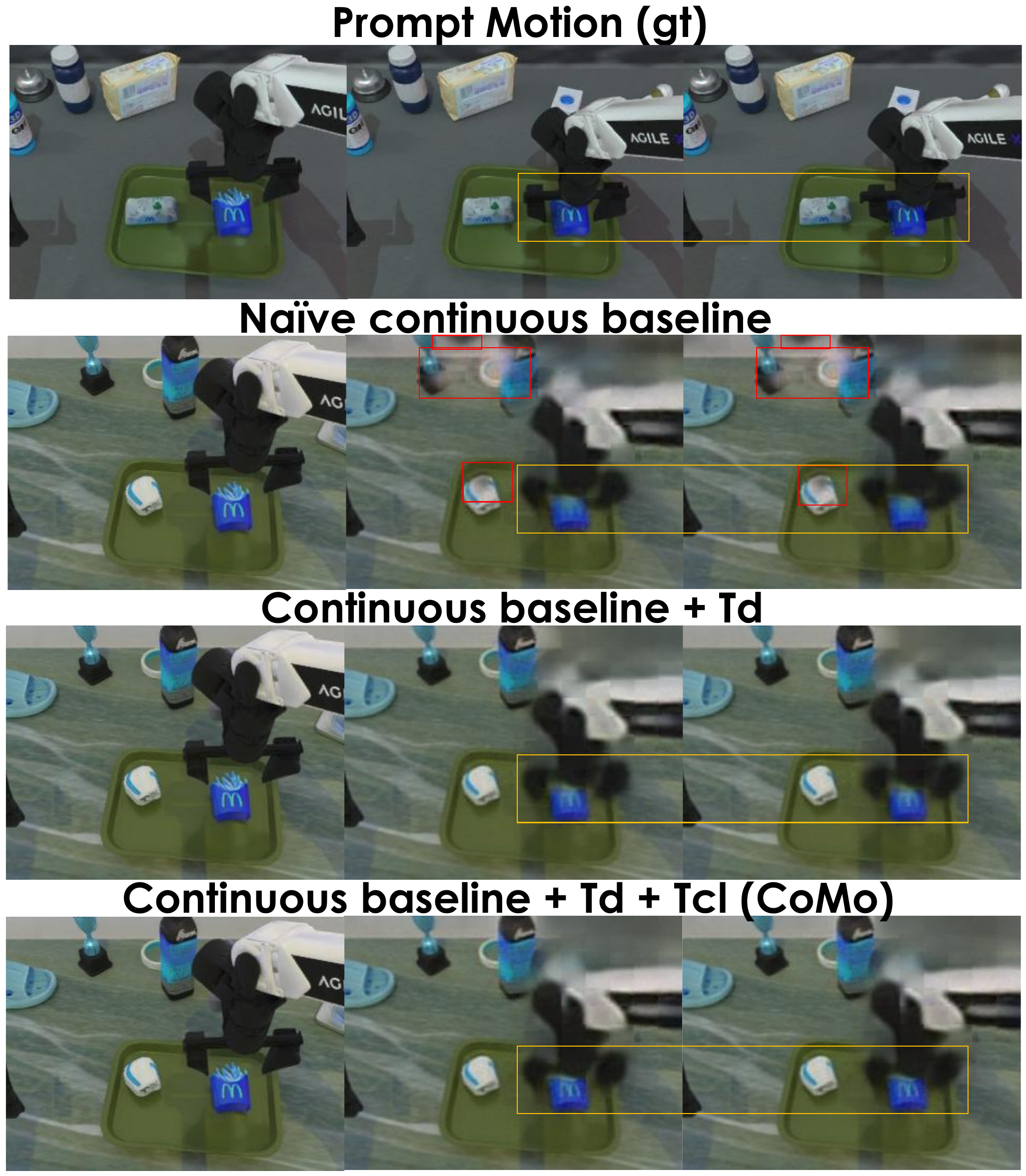} 
\caption{\textbf{FDM future frame prediction visualization.} Given sampled three frames from a prompt video, we extract latent motions from the first two and first to last frames, respectively. These motions are then used to predict the subsequent two frames via FDM in a new environment. The \textcolor{red}{red rectangles} indicate that the naïve continuous baseline significantly incorporates static background noise from the prompt video. In contrast, the early temporal difference mechanism effectively avoids this issue. Crucially, as indicated by the \textcolor{orange}{orange rectangles}, further introducing temporal contrastive learning leads to more precise latent motion representations. The results more accurately align with the fine-grained foreground motions (from non-grasp to grasp) in the prompt video. More visualizations are provided in the supplementary material.}
\vspace{-2.5mm}
\label{vis}
\end{figure}

\vspace{-2mm}

We perform experiments using the LIBERO~\citep{libero}
and CALVIN~\citep{calvin} benchmarks and a Franka Emika Research 3 robot. Our experiments study the following questions:

\noindent \textbf{Q1:} Can self-supervised CoMo extract effective pseudo action labels for action-less video data and enable unified joint training with robot data to improve policy performance?

\noindent \textbf{Q2:} What are the individual contributions of the early temporal difference (Td) mechanism and the temporal contrastive learning (Tcl) scheme, and can they synergistically improve latent motion representation learning?

\noindent \textbf{Q3:} Do continuous latent motion representations extracted by CoMo effectively mitigate shortcut learning problem and outperform discrete latent motion, naive continuous baseline, and other related methods?

\noindent \textbf{Q4:} Can the dimensions of continuous latent motion embedding extracted by CoMo be easily scaled? 

\noindent \textbf{Q5:} Can MSE and S-PCFC effectively evaluate and analyze latent motion representations and reliably reflect downstream policy success rates?

\noindent \textbf{Q6:} Does CoMo maintain its effectiveness and generalize across more complex embodiments, specifically in dual-arm and humanoid platforms with absolute action space?

\begin{figure}[htbp]
\label{zhexian}
\centering
\includegraphics[width=0.48\textwidth]{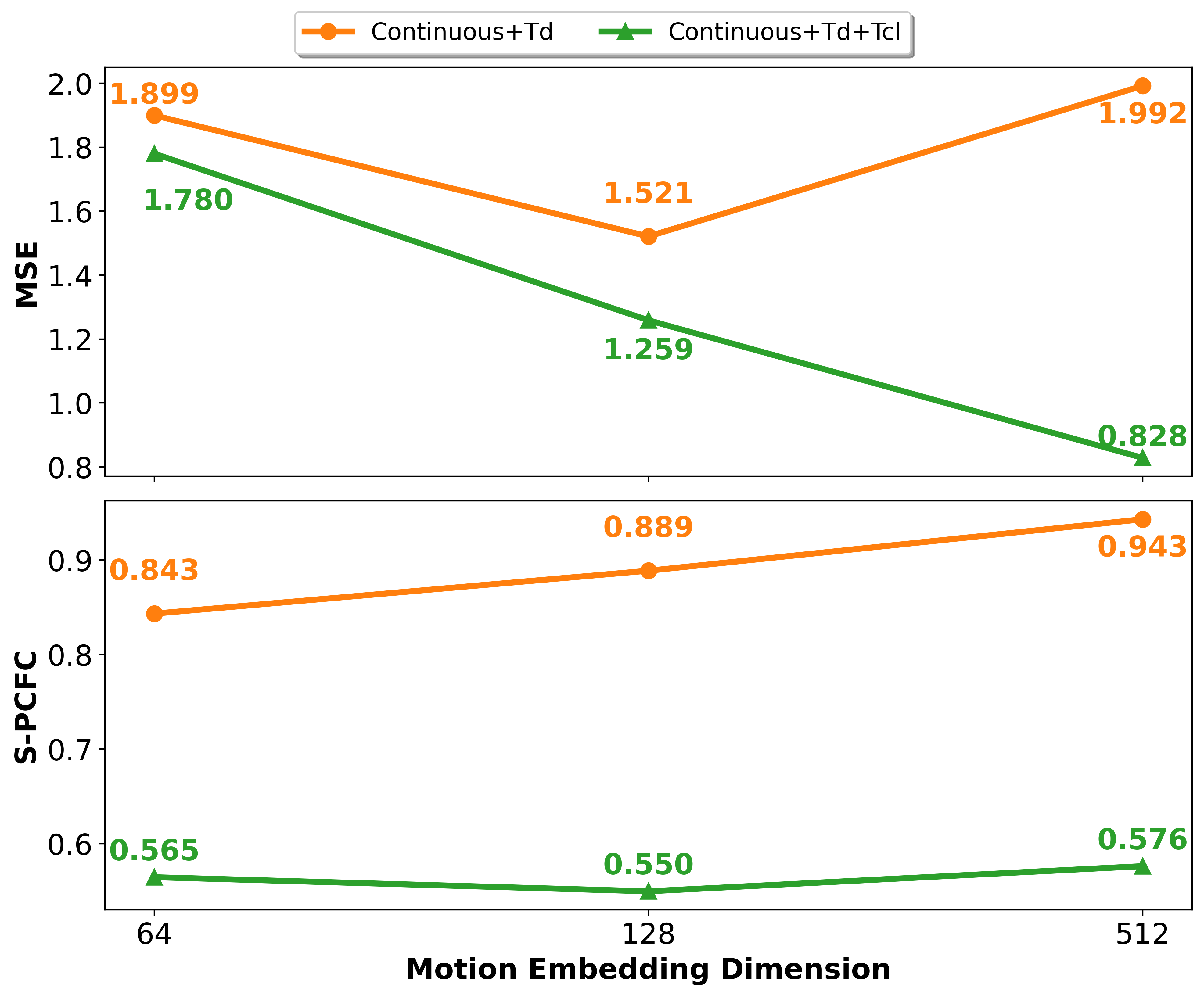} 
\caption{\textbf{Scalability of latent motion dimension in Libero.} As the latent motion dimension increases, relying solely on the temporal difference mechanism leads to a persistent increase of action-irrelevant background noise (indicated by an increasing S-PCFC). This impairs regression performance, resulting in the highest MSE at a dimension of 512. In contrast, further incorporating temporal contrastive learning (our CoMo) effectively addresses this issue and ensures the \textbf{scalability} of the latent motion dimension.}
\vspace{-4mm}
\label{zhexian}
\end{figure}

\subsection{Latent motion learning with out-of-domain Internet video data}

To evaluate the zero-shot cross-domain transfer capability of latent motion encoders, specifically, instead of training the latent motion encoders with in-domain robot data, we jointly train it on a large corpus of out-of-domain Internet videos. We uniformly sample a total of 120,000 videos from SAM-V~\citep{samv}, EgoVid~\citep{egovid}, and Droid~\citep{droid}, which cover in-the-wild, ego-centric human, and robot scenarios. Crucially, each dataset contributes 40,000 videos. All subsequent experiments utilize this exact, thereby ensuring a fair and consistent comparative analysis across all methods.

\subsection{Simulation Experiments}

\subsubsection{Simulation Benchmarks and Setups}

    \textbf{LIBERO.} The LIBERO~\citep{libero} is divided into five categories: LIBERO-Spatial, LIBERO-Object, LIBERO-Goal, LIBERO-Long, and LIBERO-90. For unified policy co-training, each task uses only 10 robot trajectories, while the remaining are video trajectories annotated with pseudo actions via latent motion IDM. Regarding the policy architecture, following prior approaches~\citep{diffusion-policy,rdt}, we implement a diffusion-based policy that conducts a joint denoising process within both the real robot action space and latent motion space. For policy evaluation, we use the final epoch model and evaluate each task with 20 trials, repeating the run three times to report the mean and standard deviation. In addition, to ensure a more robust and comprehensive ablation and analysis of the latent motion representation learning, we further report the MSE and S-PCFC results spanning 40 tasks across the four LIBERO suites.

\noindent \textbf{CALVIN.} The CALVIN~\citep{calvin} benchmark is built upon the Franka robot and focuses on assessing long-horizon task completion. It consists of four different environments (A, B, C, D), allowing for robust evaluation of generalization capabilities. We conduct experiments under the most challenging ABC $\rightarrow$ D setup, training on environments A, B, and C, and evaluating on D. Different from the setup of Moto-GPT~\citep{moto}, we use purely out-of-domain videos to train latent motion IDM and 35\% of A, B, C data (18k trajectory videos) with language annotations to conduct unified policy joint training. In terms of policy architecture, to enable joint prediction of continuous robot action and latent motion under a unified auto-regressive based policy framework, following ~\citep{moto, vla_improve}, we add two additional MLP networks after the final hidden states of the auto-regressive decoder to jointly predict robot action and latent motion.

\begin{table*}[t]
\caption{~\textbf{Ablation results on the LIBERO benchmark.} Notably, our results are achieved under a severe data constraint, utilizing \textbf{only 10} action-annotated robot demonstrations per task, a significant reduction compared to the 50 used in typical settings.}
\label{libero}
\centering
% \tablestyle{9.5pt}{1.05}
\begin{tabular}{l|c|ccccccc}
\toprule
\textbf{} & Metric & Future features & Dis. & Con. & Con.+Td & Con.+Tcl   & Con.+Td+Tcl (\textbf{CoMo})\\
\hline
\multirow{3}{*}{Spatial} & Success Rate~$\uparrow$  &76.0 $\pm$2.2 & 80.7$\pm$0.8 & 75.3$\pm$1.0 & 77.7$\pm$3.3  &80.3 $\pm$3.7   & 80.3$\pm$2.2 \\
                        & MSE~$\downarrow$ &2.1765  &5.8048 &   1.6203 &1.4811 & 1.3090  &           1.2819\\
                        & S-PCFC~$\downarrow$   &1.0000  & -0.0501 & 0.9245 & 0.8940 & 0.6658  &   0.5962  \\\midrule
\multirow{3}{*}{Object}  & Success Rate~$\uparrow$  &92.0 $\pm$0.5 &  83.3$\pm$4.1&  91.3$\pm$2.4& 95.3$\pm$5.1  &$94.3\pm$2.0   & 97.0$\pm$1.6\\
                        & MSE~$\downarrow$ & 2.0441 &  4.6786& 1.4565 & 1.1772 &1.0470   &0.9578\\
                        & S-PCFC~$\downarrow$   & 1.0000  &  -0.1521&0.9448  &  0.9075& 0.6307 & 0.6191\\\midrule
\multirow{3}{*}{Goal}    & Success Rate~$\uparrow$  & 73.3$\pm$1.7 &80.0$\pm$2.5 &  76.3$\pm$3.3& 78.7$\pm$0.9 & 77.7$\pm$0.0 & 81.0$\pm$2.4  \\
                        & MSE~$\downarrow$ & 2.1290 & 6.1323& 1.6853 &1.7340 & 1.3556  &1.3219 \\
                        & S-PCFC~$\downarrow$   & 1.0000  & -0.0649 & 0.9256 &0.8854 & 0.6316 & 0.5306\\\midrule
\multirow{3}{*}{Long}    & Success Rate~$\uparrow$  &53.7$\pm$6.4  &59.7$\pm$0.9  & 57.7$\pm$4.6& 55.7$\pm$2.5 &$59.3\pm$0.8   &62.0$\pm$3.0 \\
                        & MSE~$\downarrow$ & 2.2273 &6.0813 &1.7404  & 1.6913 &  1.5098 & 1.4737\\
                        & S-PCFC~$\downarrow$   &1.0000   &-0.0040  & 0.9124 & 0.8679& 0.5635 & 0.4523  \\\midrule
\multirow{3}{*}{Avg.}    & Success Rate~$\uparrow$  & 73.8& 75.9 & 75.2 &76.9 &  77.9&  \textbf{80.1}\\
                        & MSE~$\downarrow$ & 2.1442 & 5.6743 & 1.6256 & 1.5209& 1.3054  & \textbf{1.2588} \\
                        & S-PCFC~$\downarrow$   &1.0000 &\textbf{-0.0678}  & 0.9268 &  0.8887& 0.6229 & 0.5496   \\
\bottomrule
\end{tabular}

\end{table*}

\subsubsection{MSE and S-PCFC}

Success rate is susceptible to external factors, incurs high cost, and does not directly measure latent motion attributes. We thus adopt two complementary metrics for direct, stable, and affordable latent motion analysis and evaluation.

Action Prediction MSE (MSE). Following popular motion representation evaluation protocols~\cite{villa, clam, supervision}, we train an MLP to regress ground-truth robot actions ($a_t$) from the offline-extracted motion embeddings ($z_t$) and report the resulting MSE. MSE effectively assesses the latent motion's encoding capacity for action-relevant information.

Past-to-Current and Future-to-Current Similarity (S-PCFC). To directly measure action-irrelevant background noise and enable more fair latent motion cross-dimensional comparison, we utilize S-PCFC. This metric measures the cosine similarity between the past-to-current motion $z(o_{t-n}, o_t)$ and the future-to-current motion $z(o_{t+n}, o_t)$, where $n$ is a small time step offset. $$\mathrm{S\text{-}PCFC}(t) = \frac{z(o_{t-n}, o_t)^\top z(o_{t+n}, o_t)}{\|z(o_{t-n}, o_t)\|_2 \; \|z(o_{t+n}, o_t)\|_2}.$$ 
S-PCFC is statistically computed on the motion-centric robot dataset, which features controlled environments with minimized dynamic backgrounds, large motion and periodic movements, ensuring the metric is purely sensitive to motion dynamics and not extrinsic video noise. Since high-dimensional latent motion inevitably introduces redundant noise, a relatively lower S-PCFC indicates better motion fidelity and less static redundancy. Overall, the combination of low MSE and low S-PCFC strongly correlates with the highest downstream policy success rates, as demonstrated in our subsequent analysis.

\subsubsection{ Experiments Results and Analysis}

To comprehensively explore different latent motion learning methods, we design and implement several primary ablation variants: \textbf{Future features} (using the future frame MAE ViT features as latent motion), \textbf{Con.} (continuous latent motion baseline by naively removing vector quantization of prior works~\cite{lapas,moto}), \textbf{Dis.} (discrete latent motion by applying vector quantization, following prior works~\cite{lapas,moto}), \textbf{Con.+Td} (continuous baseline augmented only with early temporal difference), \textbf{Con.+Tcl} (continuous baseline augmented only with temporal contrastive learning), and \textbf{Con.+Td+Tcl} (continuous baseline augmented with early temporal difference and  temporal contrastive learning, i.e., our CoMo). As for `Dis.', considering the assumption of continuous data distribution in diffusion~\citep{diffusion,flowmatching1,flowmatching2}, and following GR00T~\citep{gr00t}, we utilize pre-quantized embeddings as the latent motion in diffusion-based policy. To ensure fair comparison, all the above variants use nearly identical architectures and parameter counts.

In Tab.~\ref{libero}, we present ablation studies of the above variants on LIBERO, reporting Success Rate (S.R.), MSE, and S-PCFC. In Tab.~\ref{comparsion}, we further compare the policy performance of CoMo with other related methods on LIBERO. In Tab.~\ref{calvin}, we report ablation results of the core design of CoMo on CALVIN. 
Considering that both LIBERO and CALVIN are based on single-arm with relative action space, we further apply CoMo to more complex embodiments with absolute action space, including dual-arm~\cite{robotwin} and humanoid~\cite{egovla} robots. The results are shown in Tab.~\ref{abs}. Overall, these results address the aforementioned questions and make following findings:

\noindent \textcolor{SteelBlue}{\textbf{\textit{Result 1:}}} The results in Tab.~\ref{comparsion} indicate that incorporating video data with CoMo pseudo labels into the diffusion-based policy can increase the success rate on LIBERO from 70.4\% to 80.1\%. Meanwhile, the results in Tab.~\ref{calvin} show that CoMo joint learning also improves CALVIN's performance of auto-regressive policy from 1.878 to 3.070.

\noindent \textcolor{SteelBlue}{\textbf{\textit{Finding 1 (for Q1):}}} CoMo trained on large-scale internet videos exhibits strong generalization capabilities, enabling direct zero-shot transfer to simulated robotic scenarios. CoMo can provide effective pseudo action labels for action-less video data. The ability of incorporating much richer data source and supervision leads to a more powerful policy performance, making CoMo a more scalable and data-efficient learning paradigm. 

\noindent \textcolor{SteelBlue}{\textbf{\textit{Result 2:}}} In Tab.~\ref{comparsion}, CoMo achieves the best policy performance. GR2-like and GR00T correspond to `Future features' and `Dis.', respectively. Dynamo utilizes a covariance regularization loss to suppress shortcut learning. Specifically, we incorporate the core designs of the compared methods into our unified diffusion policy. The same model architecture and training data ensures the fairness of the comparison.

\begin{table}[t]
    \caption{\textbf{Experiment results on CALVIN.} $\times 4$ indicates that we expand the latent motion dimension from 128 to 512. }
\label{calvin}
\tablestyle{2.35pt}{1.0}
% \resizebox{\textwidth}{!}{
    \centering
    \begin{tabular}{l|ccccc|c}
    \toprule
                & 1                    & 2                    & 3                    & 4                    & 5                    & Avg.            \\ \midrule
    w/o. Motion &0.772     & 0.494   & 0.307   &   0.191  &   0.114  & 1.878    \\ \midrule
         Dis.(Moto~\cite{moto})    & 0.801    & 0.575    & 0.409    & 0.283    & 0.187   & 2.255    \\ \midrule
    Con.     & 0.853    & 0.677    & 0.523    & 0.425    & 0.319    & 2.797    \\
    +Tcl    & 0.845   & 0.695    & 0.569    & 0.456    & 0.360    & 2.925    \\
    \rowcolor{blue!5}
    +Tcl+Td    & 0.882    & 0.732   & 0.589    & 0.490    & 0.377    & 3.070    \\ 
    \rowcolor{blue!5}
    +Tcl+Td ($\times 4$)    & 0.891    & 0.758   & 0.646    & 0.529   & 0.423    & \textbf{3.247} \\ \bottomrule
    \end{tabular}
% }
\end{table}

% Furthermore, the results in Tab.~\ref{real_data} demonstrate that this conclusion remains valid when CoMo is trained on larger-scale, out-of-domain Internet videos, achieving an even higher policy success rate (from 80.8\% to 81.8\%). 

\noindent \textcolor{SteelBlue}{\textbf{\textit{Finding 2 (for Q3):}}} In the comparison of different predictive signals from video data to improve robot policy, the CoMo latent motion outperforms future frame visual features, 2D point trajectory of ATM, pre-quantized latent motion of GR00T, and regularized latent motion of dynamo.

\noindent \textcolor{SteelBlue}{\textbf{\textit{Result 3:}}} We emphasize the following results: \textcolor{SteelBlue}{\textbf{(\romannumeral 1)}} Policy Success Rate (S.R.): As shown in Tab.~\ref{libero}, for diffusion-based policy, CoMo achieves an average success rate of 80.1\%, consistently outperforming all other ablation variants. In particular, it surpasses `Dis.' by a notable margin of 4.2 (increasing from 75.9\% to 80.1\%). Meanwhile, as shown in Tab.~\ref{calvin}, CoMo also attains the highest performance within the auto-regressive based policy. Compared to using discrete latent motion, the results improve from 2.255 to 3.070. \textcolor{SteelBlue}{\textbf{(\romannumeral 2)}} MSE: The results in Tab.~\ref{libero} also indicate that CoMo achieves the best performance on the MSE metric. \textcolor{SteelBlue}{\textbf{(\romannumeral 3)}} S-PCFC: For S-PCFC, the results in Tab.~\ref{libero} show that discrete latent motion achieves the lowest value. For continuous approaches, `Con.' only achieves a result close to 1.0 (0.9268). When employing early temporal difference mechanism and temporal contrastive learning, S-PCFC is significantly reduced. In summary, Td and tcl respectively contribute to the reduction of MSE and S-PCFC, as well as the improvement of S.R. The combination of Td and tcl achieves the best performance in terms of MSE, S-PCFC, and policy S.R.

\noindent \textcolor{SteelBlue}{\textbf{\textit{Finding 3 (for Q2 and Q3):}}}  Based on the above results, we have the following findings and analysis: \textcolor{SteelBlue}{\textbf{(\romannumeral 1)}} Although extracting discrete latent motion and explicitly imposing vector quantization constraints can effectively mitigate the shortcut learning problem, this approach leads to significant information loss. In addition, continuous latent motion and robot action share a consistent continuous distribution, and this consistency is beneficial for the joint learning of a unified policy. \textcolor{SteelBlue}{\textbf{(\romannumeral 2)}} Simply removing the vector quantization leads to a severe shortcut learning problem, where the model tends to collapse by directly learning substantial future frame background noise as continuous latent motion. In contrast, employing early temporal difference mechanism and temporal contrastive learning can effectively alleviate this issue (Both MSE and S-PCFC effectively reduce). More importantly, the combination of the two yields even better results through their collaborative effects. Fig.~\ref{vis} also provides supporting evidence: the latent motion from the naïve continuous baseline contains background noise from the prompt video, which is propagated to future frame prediction of new environment via FDM. Td effectively addresses this issue. However, the latent motion extracted solely by Td may be sparse. Introducing Tcl further enables the latent motion to better focus on the foreground and become more structured, thereby resulting in more accurate inter-frame changes and future frame predictions.

\begin{table}[t]
\caption{~\textbf{Comparison results with other related methods on the LIBERO benchmark.} Similarly, only \textbf{10} robot action demonstrations per task are used.}
\label{comparsion}
\tablestyle{4.0pt}{1.0}
\centering
\vspace{-0.5em}
\begin{tabular}{l|cccc|c}
\toprule
         & Spatial & Object & Goal  & Long & Avg.\\\midrule
% DP (5$\times$ data)  &92.0   &96.3     &  93.0 &75.3  &89.2 \\
DP (w/o. videos)   &  72.3 &  82.3    &  70.3 &  56.7 &70.4 \\ \midrule
ATM$^*$~\citep{atm} & 79.0   & 81.0    & 58.7& 44.0  &65.7 \\ 
GR2-like~\citep{gr2} & 76.0   & 92.0    &73.3   & 53.7 &73.8 \\

GR00T-like~\citep{gr00t} & 80.7   & 83.3   &80.0   & 59.7 &75.9 \\

Dynamo~\citep{dynamo} & 75.3   & 92.7   &80.7   & 46.0 &73.7

\\     
\rowcolor{blue!5}
CoMo & 80.3   & 97.0    & 81.0 &  62.0 & \textbf{80.1}\\ \bottomrule
\end{tabular}
\end{table}

\noindent \textcolor{SteelBlue}{\textbf{\textit{Result 4:}}} As shown in Fig.~\ref{zhexian}, when continuously increasing the latent motion dimension, relying solely on Td results in a decrease in MSE as the dimension becomes excessively high, while S-PCFC gradually rises to a relatively high value. This is because naïvely expanding the dimension undermines the information bottleneck imposed by the temporal differencing mechanism, leading to learned latent motions that are sparse and contain a substantial amount of action-irrelevant noise.

\noindent \textcolor{SteelBlue}{\textbf{\textit{Finding 4 (for Q4):}}} In contrast, further incorporating Tcl (i.e., our CoMo) avoids this issue: as the motion dimension increases, MSE continues to decrease while S-PCFC remains consistently low level. This demonstrates that CoMo (applying both Td and Tcl) ensures the latent motion focuses more on the foreground and becomes more structured, thus making its dimension more easily scalable. As shown in Tab.~\ref{calvin}, using higher-dimensional latent motion leads to higher policy success rates (3.247 vs. 3.070). This scalable property is critical for applying CoMo to a wide variety of complex embodiments and manipulation tasks.

\begin{figure*}[htbp]
\label{real_task}
\centering
\includegraphics[width=0.9995\textwidth]{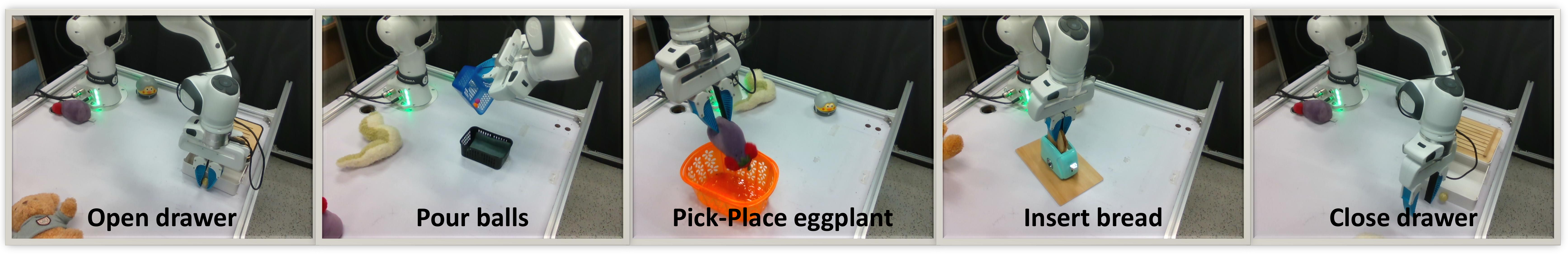} 
\caption{\textbf{Real-world task illustrations.}}
\label{real_task}
\vspace{-3mm}
\end{figure*}

\noindent \textcolor{SteelBlue}{\textbf{\textit{Finding 5 (for Q5):}}} In Tab.~\ref{libero}, we present policy S.R., MSE, and S-PCFC on LIBERO across various methods. Overall, reducing MSE or S-PCFC alone does not guarantee better policy performance. For example, `Dis.' achieves the best S-PCFC, but does not exhibit a clear advantage in success rate, especially in fine-grained tasks such as LIBERO-object (picking up smaller objects). Although `Future features' achieves a much better MSE than `Dis.', it has the lowest success rate. The best policy performance is achieved when both are relatively low, indicating a better trade-off. In summary, the combined MSE and S-PCFC are highly correlated with the downstream policy success rate. Specifically, for all continuous latent motion approaches, the average rankings of MSE, S-PCFC, and policy S.R. are consistent: \textbf{$Con. < Con.+Td < Con.+Tcl < Con.+Td+Tcl (CoMo)$}.

\begin{table}[t]
\tablestyle{1.75pt}{1.0}
\caption{\textbf{Ablation results of action prediction MSE} in absolute action space of more complex embodiments.}
\label{abs}
\centering
\begin{tabular}{ccccc}
\toprule
         & Dis. & Con. & Con.+Tcl & Con.+Tcl+Td \cellcolor{blue!5} \\ \hline
Dual-arm & 10.7611     & 6.5603     &  6.1436        & \cellcolor{blue!5} \textbf{4.9665}            \\
Humanoid & 0.3573     & 0.0889     & 0.0789         &  \cellcolor{blue!5} \textbf{0.0732}          \\ 
\bottomrule
         &      &      &          &            
\end{tabular}
 \vspace{-2.5mm}
\end{table}

 \noindent \textcolor{SteelBlue}{\textbf{\textit{Finding 6 (for Q6):}}} We primarily conduct simulation ablation experiments on LIBERO and CALVIN, both of which employ Frank single-arm robots and a 7-dimensional, relative action space. To further validate the generalizability of our proposed Td and Tcl, we report MSE ablation results on dual-arm and humanoid (equipped with dexterous hands) robots, which both use absolute action space with 14 and 128 dimensions, respectively. The results in Tab.~\ref{abs} further reinforce the effectiveness of our proposed Td and Tcl method, demonstrating its generalizability.

% \vspace{-0.5mm}

\subsection{Real-world Experiments}

% \vspace{-0.5mm}

 \noindent \textbf{Real-World Setups.} In real-world experiments, we aim to validate whether CoMo trained on Internet videos, can directly extract latent motion for human hand videos to serve as more effective pseudo labels. Specifically, we utilize a Franka robot to execute five tasks: picking up a toy and placing it into the basket, opening the drawer, closing the drawer, inserting the bread into the container, and pouring the balls into the basket, as shown in Fig.~\ref{real_task}. For each task, we utilize 25 teleoperated trajectories and 25 human hand manipulation videos on the same environments, where we construct pseudo action labels via latent motion IDM for human hand videos. We then train a unified diffusion-based policy using both datasets and evaluate each task with 20 rollouts, and the results are presented in Tab.~\ref{real_result}.

 \noindent \textbf{Real-World Results and Analysis.} The results in Tab.~\ref{real_result} demonstrate the effectiveness of CoMo latent motion in extracting pseudo-action labels from human demonstration videos. Consistent with the conclusions from simulation experiments, CoMo latent motion achieves the best policy performance than naïve discrete or continuous baseline, which can be attributed to its continuous distribution that matches real robot action, more precise latent motion capture, and reduced action-irrelevant background noise. Notably, this advantage is especially pronounced in low-tolerance tasks that require fine manipulation, such as opening the drawer and inserting the bread.

\begin{table}[t]
  \tablestyle{1.75pt}{1.0}
  \caption{~\textbf{Real-world experiment results on Franka robot arm.}}
  \label{real_result}
  \centering
  \begin{tabular}{c|ccccc}
    \toprule
    & Pick-Place & Open & Insert & Close & Pour \\
    \hline
    w/o. human motion &   55.0      &   35.0      &   10.0   &   80.0  &   45.0 \\ \hline
    Discrete         &    \textbf{70.0}     &  40.0       &      10.0    &   85.0  &   \textbf{55.0}  \\
    Continuous         &    65.0     &  45.0       &      25.0    &   \textbf{90.0}  &   50.0  \\
    \rowcolor{blue!5}
    CoMo (+Td+Tcl)            &  \textbf{70.0}       &  \textbf{60.0}       &     \textbf{35.0}     &   \textbf{90.0}  &   \textbf{55.0} \\
    \bottomrule
  \end{tabular}
  \vspace{-2.5mm}
\end{table}

\vspace{-1mm}

\section{Conclusions}
% \vspace{-0.5mm}
We presented CoMo, a self-supervised framework of learning continuous latent motion from Internet videos. By employing the early temporal difference and temporal contrastive learning methods, CoMo effectively mitigates shortcut learning issues, thus enabling more effective and precise pseudo action labels for action-less video data. This seamlessly facilitates the joint learning of continuous robot action and latent motion within a unified policy. Furthermore, we propose MSE and S-PCFC for a more direct and stable evaluation and analysis of latent motion, revealing superior properties of the latent motion learned by CoMo.

% \vspace{-1mm}

\section{Limitations and Discussion}

Extensive simulation and real-world experiments validate CoMo’s effectiveness and generalizability in extracting pseudo labels for action-less videos and demonstrate its superiority over discrete latent motion and naïve continuous baselines. Despite CoMo yielding vastly superior MSE and S-PCFC metrics, we note that policy co-training gains remain modest in certain tasks. We attribute this to CoMo being fundamentally more advantageous for fine-grained tasks. Therefore, constructing larger-scale and more complex benchmarks for real-world fine-grained manipulation tasks will be crucial. Additionally, extending CoMo to multi-frame and multi-view modeling represents a highly valuable direction to mitigate occlusions and push the boundaries of motion extraction accuracy.

\noindent \textbf{Acknowledgements.}
This work is supported by the National Key R\&D Program of China (No. 2022ZD0160900), the Basic Research Program of Jiangsu (No. BK20250009), the Fundamental and Interdisciplinary Disciplines Breakthrough Plan of the Ministry of Education of China (No. JYB2025XDXM118), and the Collaborative Innovation Center of Novel Software Technology and Industrialization.

\setcounter{section}{0}

\renewcommand\thesection{\Alph{section}}

\section{Appendix}

\subsection{Real-World Experiments Details}
\label{real}
\begin{figure*}[htbp]
\centering
\includegraphics[width=0.985\textwidth]{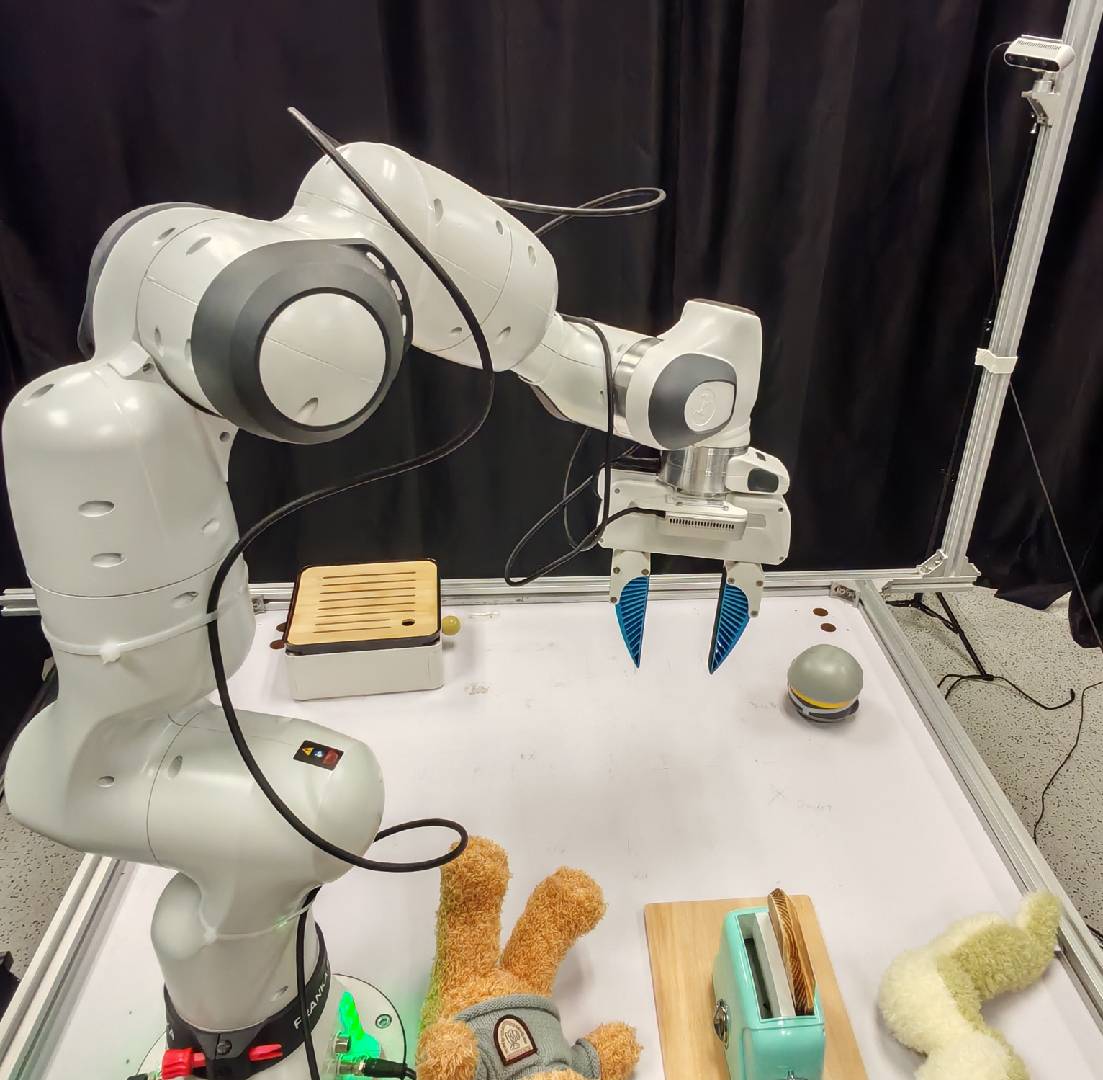} % Reduce the figure size so that it is slightly narrower than the column.
\caption{\textbf{The real-world Franka robot arm experiments hardware platform.}}
\label{setups}
\end{figure*}

In this section, we detail the specifics of our real-world experiments. Specifically, our experiments setup is illustrated in Fig.~\ref{setups}, which comprises a single Franka Emika Research 3 robot arm, equipped with a UMI~\cite{umi} gripper, and utilizes a statically positioned RealSense D435 camera (with a resolution of 640×480 pixels) from a third-person view to acquire real-time RGB visual observations. Following publicly available code\footnote{\url{https://github.com/UT-Austin-RPL/deoxys_control}}, we employ a 3D mouse for teleoperation data collection. The robot system operates at 20 Hz (moderately reduced from the native 100 Hz control frequency to balance training efficiency and motion continuity), with actions defined as relative end-effector pose changes in SE(3) space (3D position change + quaternion orientation change + gripper state).

For five real-world tasks we evaluated—picking up corresponding toy and placing it into the basket, opening the drawer, closing the drawer, inserting the bread into the container, and pouring the balls into the basket—they respectively require the robot arm to perform basic picking-and-placing, fine-grained and contact-rich opening, contact-rich closing, fine-grained picking-inserting, and picking-pouring capability. During evaluation, the initial pose of the robot arm was set to a fixed home position. The initial poses of the objects to be interacted with were significantly varied. A special case is the opening-drawer and closing-drawer task, where adhesive was applied to the bottom of the drawer to mitigate significant sliding during opening and closing. Consequently, in this task, the placement pose of the drawer was slightly perturbed, within a range of approximately 8 cm in the lateral and longitudinal directions. 

As for the policy of our real-world experiments, we adopt a diffusion-based policy architecture. Specific training and architecture details can be found in Section~\ref{diffusion}. Finally, we jointly train the policy using collected robot data and human hand video data labeled with the corresponding latent motion IDM.

% and present several real-world execution demonstrations in Fig.~\ref{demos}.

\subsection{ CoMo Details}
\label{como}

In this section, we describe the specifics of our CoMo. As shown in Tab.~\ref{como_hyperparams}, we report the training and architectural details of our CoMo. We aim to learn a generalizable latent motion IDM that can extract latent motion representing any form of inter-frame changes. To this end, we uniformly sample a total of 120,000 videos from SAM-V~\cite{samv}, EgoVid~\cite{egovid}, and Droid~\cite{droid}, with each dataset contributing 40,000 videos. These datasets collectively cover both ego-centric and fixed-camera viewpoints, and encompass a wide range of motions, including those of robotic arms, humans, and various objects in the wild. Importantly, all of our baselines employ the same model architecture, training data, and hyperparameters as CoMo, which ensures the strict fairness of our comparisons.

Specifically, for the discrete latent motion baseline, there is a trade-off regarding the choice of codebook size. A larger codebook size typically enables more comprehensive capture of motion information, but also increases the risk of encoding action-irrelevant background noise. Conversely, a smaller codebook size may limit the captured motion details but reduces such noise. In our experiments with the discrete latent motion baseline, we compare a codebook size of 8 (following LAPA~\cite{lapas}) in the LIBERO and real-world settings, and a codebook size of 128 (following Moto-GPT~\cite{moto}) in CALVIN. In all cases, the results consistently demonstrate the superiority of our CoMo.

\begin{figure*}[htbp]
\centering
\includegraphics[width=0.9985\textwidth]{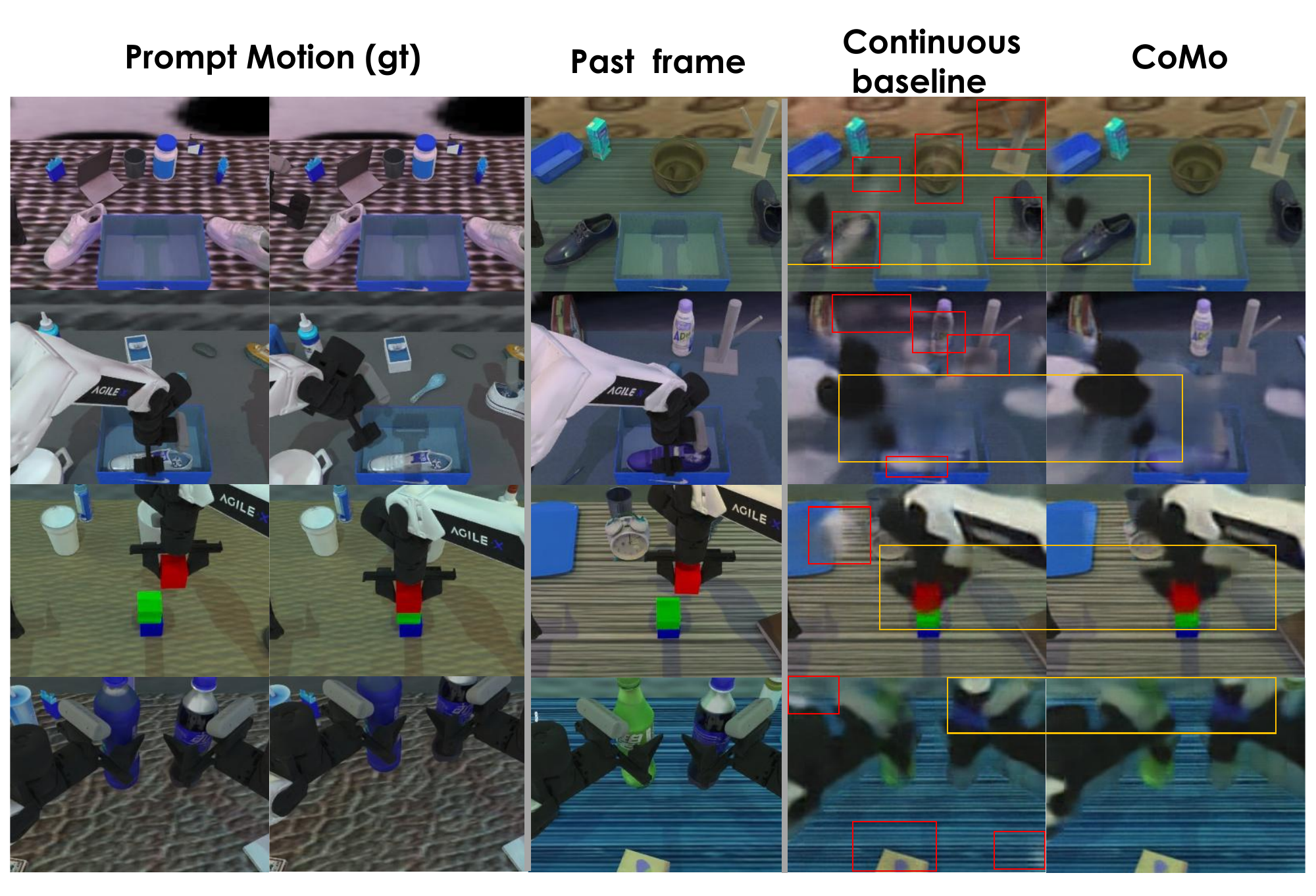} % Reduce the figure size so that it is slightly narrower than the column.
\caption{\textbf{The FDM future frame prediction visualization.}}
\label{viss}
\end{figure*}

\begin{table*}[htbp]
 \caption{The training and architectural hyperparameters for our CoMo learning.}
    \centering
    \setlength{\tabcolsep}{6.0pt}
    \begin{tabular}{ll}
        \toprule
        \textbf{Hyperparameter} &  \textbf{Value}\\\hline
        \multicolumn{2}{c}{\textit{CoMo training}}\\\hline
        Optimizer & AdamW~\cite{adam} \\
        Base learning rate & 0.0001 \\
        % weight decay & 0.05 \\
        Optimizer momentum & $\beta_1, \beta_2=0.9, 0.99$\\
        Effective batch size & 256 \\
        % learning rate schedule & cosine decay \\
        Total training steps & 50,000\\
        % Frame interval on LIBERO~\cite{libero} & 10\\
        % Frame interval on CALVIN~\cite{calvin} & 5\\
        Frame interval on SAM-V~\cite{samv} & 10 \\
        Frame interval on EgoVid~\cite{egovid} & 10\\
        Frame interval on Droid~\cite{droid} & 20\\
        % warmup epochs & 5 \\
        % augmentation & {RandomResizedCrop (0.8, 1)} \\
        \hline
        \multicolumn{2}{c}{\textit{Inverse dynamics Model}}\\
        \hline
        Feature extractor & MAE~\cite{mae} ViT-L\\
        Codebook size of discrete baseline & 8 and 128\\
         Number of motion queries & 8\\
         Latent motion embedding dimensionality  & 16 \\
        \#layers & 4\\
        \#MHSA heads & 12\\
        Hidden dim & 768\\
        % positional embedding & sin-cos initialization and fix\\
        \hline
        \multicolumn{2}{c}{\textit{Forward dynamics Model}}\\
        \hline
        \#layers & 12\\
        \#MHSA heads & 12\\
        Hidden dim & 768\\
        % positional embedding & sin-cos initialization and fix\\
        \bottomrule
    \end{tabular}
    % }
     \label{como_hyperparams}
\end{table*}

\subsection{Diffusion-based Policy Details}
\label{diffusion}

In this section, we detail our unified diffusion-based policy. We primarily implement the diffusion-based policy for the LIBERO~\cite{libero} simulation and real-world experiments. Specifically, we jointly learn the unified diffusion-based policy from video data with pseudo action labels constructed using the corresponding latent motion IDM, and continuous robot action data. 

In Tab.~\ref{dp_hyperparams}, we report the training and architectural details of our diffusion-based policy. Specifically, we employ BERT~\cite{bert} and ViT~\cite{vit} to extract language instructions and visual observations features, respectively. Following RDT-1B~\cite{rdt}, we utilize a more scalable DiT~\cite{dit} block as the backbone. The extracted language and visual features are incorporated as conditioning through cross-attention layers within the DiT block. To perform joint learning of action-less video data and robot data within a unified policy model, we construct two sets of MLP networks to map latent motion and robot actions into a shared embedding space, and back to their respective original spaces. In the training phase, we adopt the DDPM scheduler with a glide cosine scheduling scheme (specifically, the squaredcos cap v2 variant) across a diffusion process of 1000 steps. Conversely, for inference, we leverage the DPM-Solver++~\cite{dpm} in conjunction with an analogous glide cosine scheduler, albeit with a substantially reduced sampling budget of 5 steps. Finally, to capture the temporal dependencies of actions and ensure real-time dynamic adaptability during policy execution, we set an action / motion chunk size of 8 in both the training and inference phases.

\begin{table*}[htbp]
 \caption{The training and architectural hyperparameters for our diffusion-based policy learning.}
    \centering
    \setlength{\tabcolsep}{4.0pt}
    \begin{tabular}{ll}
        \toprule
        \textbf{Hyperparameter} &  \textbf{Value}\\\hline
        \multicolumn{2}{c}{\textit{Diffusion-based policy training}}\\\hline
        Optimizer & AdamW~\cite{adam} \\
        Base learning rate & 0.0005 \\
        % weight decay & 0.05 \\
        % Optimizer momentum & $\beta_1, \beta_2=0.9, 0.99$\\
        Effective batch size & 256 \\
        % learning rate schedule & cosine decay \\
        Total training epochs & 100\\
        \hline
        \multicolumn{2}{c}{\textit{Diffusion-based policy architecture}}\\
        \hline
        Vision feature extractor & DINOv2~\cite{dinov2} ViT-B~\cite{vit}\\
        Language feature extractor &  BERT~\cite{bert}\\
        \#layers & 12\\
        \#MHSA heads & 16\\
        Hidden dim & 768\\
        Action / motion chunk size & 8\\
        Action projector  & (7, 768)\\
        Latent motion projector  & (128, 768)\\
        Action head  & (768, 7)\\
        Latent motion head  & (768, 128)\\
        % positional embedding & sin-cos initialization and fix\\
        \hline
        \multicolumn{2}{c}{\textit{Noise scheduler}}\\
        \hline
        Type & DDPM~\cite{diffusion}\\
        Prediction type  & sample\\
        Training step number & 1000\\
        Sampling step number & 5   \\
        Solver & DPM-Solver++~\cite{dpm} \\

        % Hidden dim & 768\\
        % positional embedding & sin-cos initialization and fix\\
        \bottomrule
    \end{tabular}
    % }
     \label{dp_hyperparams}
\end{table*}

% \vspace{15mm}

\begin{table*}[htbp]
 \caption{The training and architectural hyperparameters for our auto-regressive based policy learning.}
    \centering
    \setlength{\tabcolsep}{5.5pt}
    \begin{tabular}{ll}
        \toprule
        \textbf{Hyperparameter} &  \textbf{Value}\\\hline
        \multicolumn{2}{c}{\textit{Auto-regressive based policy training}}\\\hline
        Optimizer & AdamW~\cite{adam} \\
        Base learning rate & 0.0005 \\
        weight decay & 0.0001 \\
        % Optimizer momentum & $\beta_1, \beta_2=0.9, 0.99$\\
        Effective batch size & 512 \\
        % learning rate schedule & cosine decay \\
        Total training epochs & 20\\
        \hline
        \multicolumn{2}{c}{\textit{Auto-regressive based policy architecture}}\\
        \hline
          Vision feature extractor & MAE~\cite{mae} ViT-B~\cite{vit}\\
        Language feature extractor &  T5~\cite{T5}\\
        % Vision feature extractor & DINOv2~\cite{dinov2}-pretrained ViT-Base~\cite{vit}\\
        % Language feature extractor &  BERT~\cite{bert}\\
        \#layers & 12\\
        \#MHSA heads & 12\\
        Hidden dim & 768\\
        Action chunk size & 5\\
        Motion chunk size & 2\\
        \bottomrule
    \end{tabular}
    % }
     \label{ar_hyperparams}
\end{table*}

\subsection{Auto-regressive based Policy Details}
\label{ar}

In this section, we detail the specifics of our auto-regressive based policy, as shown in Tab.~\ref{ar_hyperparams}. We primarily implement this policy for the CALVIN~\cite{calvin} simulation environment experiments. Specifically, we employ T5~\cite{T5} and ViT~\cite{vit} to extract token-level textual and visual features, respectively. Following ~\cite{moto,vla_improve}, we adopt a GPT-style~\cite{gpt1} auto-regressive backbone and append two additional MLP networks at the output layer to predict continuous robot actions and latent motion separately using the MSE loss. For the discrete latent motion baseline, in accordance with Moto-GPT~\cite{moto}, we utilize the cross-entropy loss function to optimize latent motion prediction. Additionally, we incorporate an action MLP network, consistent with the continuous approach, and employ an MSE loss to learn the robot action.

Specifically, for motion prediction, we auto-regressively predict latent motion with a chunk size of 2. For action prediction, we parallelly decode actions with a chunk size of 5 based on a set of learnable action query tokens. Furthermore, to ensure a fair comparison with the discrete baseline in Moto-GPT~\cite{moto}, we first perform a round of latent motion prediction pre-training using action-less video data before conducting joint training on robot action data and action-less video data.

\subsection{FDM future frame prediction visualization}

In this section, we present further visualizations of FDM future frame predictions to qualitatively assess the advantages of CoMo compared to the naïve continuous baseline, as shown in Fig.~\ref{viss}. Specifically, given two frames from a prompt video, we extract the latent motion between them. This extracted motion is then used to predict the subsequent frame via FDM in a new environment. The red rectangles highlight that the naïve continuous baseline tends to incorporate significant static background noise from the prompt video. In contrast, our CoMo effectively avoids this issue. Notably, as indicated by the orange rectangles, CoMo produces more precise latent motion representations, resulting in predictions that more accurately reflect the fine-grained foreground motions present in the prompt video.

\subsection{Dual-arm and humanoid MSE results details}

In the main text, we report ablation results of MSE on more complex robotic platforms, including dual-arm and humanoid robots equipped with dexterous hands. Specifically, for the dual-arm robot, we use RoboTwin~\cite{robotwin} dataet, where the action space consists of the absolute joint states of both arms (Aloha AgileX), totaling 14 dimensions. For the humanoid robot platform, we utilize the early open-source EgoVLA~\cite{egovla} dataset, which contains both human motion capture data and humanoid robot data. The EgoVLA adopts a shared action space for humans and humanoid robots, comprising the absolute wrist pose and MANO hand parameters, for a total of 128 dimensions. Overall, the experimental results underscore the effectiveness of CoMo as a unified action space for cross-embodiment data. Notably, it demonstrates robust versatility across both relative and absolute action spaces, particularly within high-dimensional contexts that require the capture of fine-grained motions.

{
    \small
    \bibliographystyle{ieeenat_fullname}
    \bibliography{main}
}


\end{document}